\documentclass[journal]{IEEEtran}
\usepackage{amsmath}
\usepackage{amssymb}
\usepackage{algorithm}
\usepackage{algorithmic}
\usepackage{comment}
\usepackage{mathtools}
\usepackage{graphicx}
\usepackage{tabularx}
\usepackage{adjustbox}
\usepackage{subcaption}
\usepackage{varwidth}
\usepackage{xcolor}
\usepackage{booktabs}
\usepackage{pgfplotstable}
\usepackage{lineno}
\usepackage{hyperref}
\modulolinenumbers[5]
\definecolor{RevCol}{RGB}{255,0,0}

\begin{document}

\title{From Distributional to Quantile Neural Basis Models: the case of Electricity Price Forecasting}

\author{Alessandro Brusaferri, Danial Ramin, and Andrea Ballarino%
\thanks{Alessandro Brusaferri (corresponding author), Danial Ramin, and Andrea Ballarino are with CNR, Institute of Intelligent Industrial Technologies and Systems for Advanced Manufacturing, via A. Corti 12, Milan, Italy (e-mail: alessandro.brusaferri@cnr.it).}%
}

\maketitle

\begin{abstract}
While neural networks are achieving high predictive accuracy in multi-horizon probabilistic forecasting, understanding the underlying mechanisms that lead to feature-conditioned outputs remains a significant challenge for forecasters.
In this work, we take a further step toward addressing this critical issue by introducing the Quantile Neural Basis Model, which incorporates the interpretability principles of Quantile Generalized Additive Models into an end-to-end neural network training framework. To this end, we leverage shared basis decomposition and weight factorization, complementing Neural Models for Location, Scale, and Shape by avoiding any parametric distributional assumptions.
We validate our approach on day-ahead electricity price forecasting, achieving predictive performance comparable to distributional and quantile regression neural networks, while offering valuable insights into model behavior through the learned nonlinear mappings from input features to output predictions across the  horizon.
\end{abstract}

\begin{IEEEkeywords}
Neural Networks, Generalized Additive Models, Quantile regression, Distributional regression, Time-series, Probabilistic Forecasting, Electricity Price
\end{IEEEkeywords}

\section{Introduction}
The challenge of probabilistic electricity price forecasting (PEPF) in day-ahead power markets constitutes a critical research area with significant practical implications. Generating accurate predictions while reliably quantifying uncertainty is essential for numerous market participants, including utilities, retailers, aggregators, and major consumers \cite{MASHLAKOV2021116405}. Electricity prices exhibit distinctive characteristics such as high volatility and abrupt fluctuations, driven by complex factors including distributed energy demand patterns, production costs, and meteorological conditions \cite{CIARRETA2022}. The increasing integration of renewable energy technologies further complicates this environment, despite their vital role in reducing global emissions \cite{MADADKHANI2024107241}. 
Over recent decades, a variety of methodologies have been developed to capture and represent the inherent uncertainty in probabilistic electricity price forecasts. These approaches range from basic prediction interval techniques to advanced methods such as discrete conditional quantile estimation and comprehensive distributional modeling \cite{BRUSAFERRI20191158}. For a detailed overview of these developments, readers are referred to recent survey articles \cite{NOWOTARSKI20181548},\cite{LAGO2021116983}.

Currently, the field witnesses growing attention toward neural network-based methodologies, which capitalize on their adaptive mapping abilities within the conditioning space while benefiting from expanding computational resources and advanced tooling \cite{LAGO2021116983}.  Within this paradigm, distributional deep neural networks (D-DNNs) that parameterize sophisticated functional forms (see e.g., \cite{MARCJASZ2023106843}) and quantile-regression deep neural networks (QR-DNN) (see e.g., \cite{brusaferri2024onlineconformalizedneuralnetworks}) have demonstrated substantial promise in modeling the intricate characteristics of price distribution patterns. These include pronounced heteroskedasticity, heavy-tailed behavior, and asymmetries. 

Nevertheless, such enhanced modeling capabilities introduce challenges regarding model interpretability. The fundamental opacity of these architectures limits understanding of the forecast generation process, potentially creating barriers to deployment in critical decision-making contexts. Most significantly, the mechanisms through which these models transform input features into density parameters and output quantiles across the forecasting horizon remain largely inaccessible to practitioners. 
Various post-hoc model-agnostic approaches, including LIME-based surrogate modeling and extensions of Shapley value methods, have been developed to provide interpretability through local model approximations and feature contribution analysis \cite{TSCHORA2022118752}.  However, achieving reliable global interpretations of the intricate internal computations executed by these models remains fundamentally challenging \cite{rudin19}. 

Consequently, an alternative research direction focuses on incorporating transparency directly into neural network design. A notable example within this category is the Neural Additive Model (NAM) framework proposed by \cite{10.5555/3540261.3540620} and subsequently applied to forecasting problems in \cite{JO2023120307}. Building upon the established Generalized Additive Models (GAM) structure, NAMs employ linear combinations of specialized neural networks, with each network processing individual features, thereby exposing the underlying input-output relationships. Beyond PEPF applications, recent work by \cite{pmlr-v238-frederik-thielmann24a} has expanded NAMs from their initial point prediction focus to encompass comprehensive distributional regression, resulting in Neural Additive Models for Location, Scale, and Shape (NAMLSS) through adaptation of the broader GAMLSS framework \cite{hirsch2024onlinedistributionalregression}. Building upon this work, the NAMLSS models has been further developed and deployed on PEPF tasks in \cite{NBMLSS} by leveraging a basis decomposition of the feature shape functions \cite{10.5555/3600270.3600882}, where a unique dense map is adopted to learn a set of shared basis combined by trainable linear projections. While achieving probabilistic forecasting performance comparable to that of D-DNNs parameterizing flexible Johnson's SU densities, such architecture (labeled NBMLSS) provides more insights into the model behavior through the learned nonlinear feature level maps to the distribution parameters across the prediction steps. 
Still, the NBMLSS model is constrained by relying on parametric assumptions, as is the case with the baseline D-DNNs. 

In this work, we further explore the compelling Neural Additive Model framework for probabilistic forecasting, with a particular focus on the challenging task of day-ahead electricity price prediction. To this end, we propose a nonparametric quantile regression counterpart to the NBMLSS architecture, inspired by the design principles of Quantile Generalized Additive Models \cite{GAILLARD20161038}, \cite{JSSv100i09} and conceived to enhance interpretability within the broader QR-DNN framework. We label the developed method QNBM hereafter, standing for Quantile Neural Basis Model.
Experimental comparisons with D-DNN, QR-DNN, and NBMLSS models are carried out using publicly accessible datasets from the day-ahead electricity markets of Germany and Belgium, encompassing recent time periods and diverse market characteristics.

\section{Methods}
\label{Methods}
We concentrate on the class of multi-step probabilistic forecasting models designed to identify a discrete set of target distribution quantiles $\gamma \in \Gamma$ (e.g., deciles, percentiles, etc.) over the entire prediction horizon $h=1,...,H$ (e.g., the next 24 hours). Still, we emphasize that the proposed approach can be readily applied to sub-cases employing distinct models to estimate different quantiles or to perform stage-specific inference. 
In this context, neural networks are typically implemented as parameterized functions mapping the input variables set to the predictions over the horizon in a unique pass:
\begin{equation}
\resizebox{\columnwidth}{!}{$
p({y}_{t+1}\ldots{y}_{t+h} \mid y_{t-k:t}, z_{t-k:t}, x_{t+h}) = f_\Theta\left(y_{t-k:t}, z_{t-k:t}, x_{t+h} \right)
$}
\label{f}
\nonumber
\end{equation}
where $\Theta$ summarizes the parameters of the model, and $k$ is the maximum lag involved in the observed history of each input series.
In PEPF applications, the conditioning variable set in the right-hand side often comprises the past values of the target price $y_{t-k:t}$, which is available till the current day, as well as further exogenous features. The exogenous set can include both observations from the past days $z_{t-k:t}$ (e.g., the previous electricity demands) and predicted variables $x_{t+h}$, such as the electricity load forecast, renewable generation predictions, as well as constant features (e.g., day-of-week encoding, etc).
Following the feed-forward class of the deep neural network (DNN) model proposed in \cite{MARCJASZ2023106843} for PEPF, the conditioning set is then structured as a flattened tensor of size $n_f$, defined as $\mathbf{x}_d=[x_{d,1},...,x_{d,n_f}]$ hereafter.
For consistency, the target price values over the prediction horizon for each $d$-th day, e.g., next 24 hours, are labeled $\mathbf{y}_d=[y_{d}^1,...,y_{d}^H]$.

\subsection{From NBMLSS to Quantile Neural Basis Models}
Formally, the Quantile Neural Basis Model architecture (QNMB) is stated as follows:
\begin{align}
z_{k}(x_{d,i})=& \mathbf{a}\left[ \sum_{j=1}^{n_u}\omega_{j,k}^{(2)} \mathbf{a} \left[\omega_{j}^{(1)}x_{d,i}\right]+\omega_{0,k}^{(2)}\right], k=1,...,n_z \label{bfe} \\
f_i(x_{d,i})=&\sum_{k=1}^{n_z}W_{(i,k)}z_{k}(x_{d,i}), i=1,...,n_f\\
\hat{q}_{h}^{{\gamma}}(\mathbf{x}_d)=&\beta_h^\gamma + \sum_{i=1}^{n_f}V_{(h,\gamma,i)}f_i(x_{d,i}), h=1,...,H, \gamma \in \Gamma \label{bout}
\end{align}
Apparently, the design follows the basis expansion framework of the NBMLSS model, but replaces the final layers - originally used to map link functions to the distribution parameters - with a direct mapping to output-conditioned quantiles.

In detail, Equation~\ref{bfe} defines the $k$-th shared basis function within a set of cardinality $n_z \in \mathbb{Z}^+$, $\omega{j}^{(1)} \in \mathbb{R}^{n_u}$ and $\omega_{j,k}^{(2)} \in \mathbb{R}^{n_u \times n_z}$ represent the trainable weights of the first and second hidden layers of the shared neural network, respectively, while $\mathbf{a}[\cdot]$ denotes the nonlinear activation functions applied throughout the network. Although the original Neural Additive Model proposed exponential activation functions to capture highly non-linear and jagged feature mappings \cite{10.5555/3540261.3540620}, recent research has demonstrated that standard ReLU activations often provide superior performance in practice \cite{bouchiat2024improvingneuraladditivemodels}.
Besides, Dropout layers are incorporated to promote decorrelation among basis functions by randomly omitting units during training.

The subsequent equations model the stepwise target conditional distribution quantiles $\gamma \in \Gamma$ through trainable linear projections derived from the basis output $z_k(x_{d,i})$, evaluated for each input feature $x_{d,i}$. Specifically, the matrix $W\in\mathbb{R}^{n_z \times n_f}$ serves to aggregate the shared basis functions into feature-specific shape functions $f_i(\cdot)$, while the tensor $V\in \mathbb{R}^{H \times |\Gamma| \times n_f}$ combines the outputs of these shape functions to produce the stage-wise quantile estimates $\hat{q}_{h}^{\gamma}(x_d)$. $\beta_h^\gamma \in \mathbb{R}$ denotes the trainable bias terms. 

Still, the straight formulation in Equations~\ref{bfe}–\ref{bout} requires not only storing the full weighting tensors $W$ and $V$, but also computing and retaining their gradients during backpropagation. As their dimensions grow rapidly with the number of features and quantiles across the prediction horizon, this results in substantial memory consumption and computational overhead during training. To mitigate these challenges, we introduce trainable low-rank approximations via matrix factorization. Specifically, for a matrix $M \in \mathbb{R}^{m \times n}$, we approximate it as:
\begin{equation}
    M \approx A B^\top, \text{  with: }A \in \mathbb{R}^{m \times r}, \; B \in \mathbb{R}^{n \times r}, \; r \ll m, n
\end{equation}
Both $W$ and $V$ can be factorized in this way, depending on the application’s requirements. The tensor $V$ is transformed into $\tilde{V} \in \mathbb{R}^{H*|\Gamma| \times n_f}$ followed by reshaping in proper form. The rank $r$ parameter controls the compression/accuracy trade-off. Besides the computational advantage, the factorization is aimed to enforce the identification of more compact and possibly robust representations \cite{hu2021loralowrankadaptationlarge}. 

The overall architecture is trained end-to-end (e.g., through the Adam optimizer) by minimizing the average multi-stage Pinball loss across the discrete set $\gamma \in \Gamma$ predicted conditional quantiles at each stage $\hat{q}_{{h}}^{\gamma}(\mathbf{x}_d)$, computed as:
\begin{multline}
    \sum_{d}{\sum_{h}}\sum_{\gamma}(y_d^{{h}}-\hat{q}_{{h}}^{\gamma}(\mathbf{x}_d))\gamma {1}\{y_d^{{h}}>\hat{q}_{{h}}^{\gamma}(\mathbf{x}_d)\} \\
+ (\hat{q}_{{h}}^{\gamma}(\mathbf{x}_d)-y_d^{{h}})(1-\gamma) {1}\{y_d^{{h}}\leq \hat{q}_{{h}}^{\gamma}(\mathbf{x}_d)\}
\end{multline}

where the datasets $\mathcal{D}_n \equiv \{(\mathbf{x}_d, \mathbf{y}_d)\}_{d=1}^n$ are generated by segmenting both the conditioned and exogenous time series using a moving window approach, and subsequently merging the resulting slices into unified input-output pairs. To prevent overfitting, a validation-based early stopping strategy is applied, where training halts if no improvement is observed over a specified number of epochs. Besides, we integrated Reversible Instance normalization (RevIN) layers (see \cite{kim2021reversible}) beyond the common model recalibration, designed to mitigate potential distribution shifts.
Further implementation aspects (such as hyperparameter tuning, etc.) are discussed in the next section dedicated to the experimental evaluation.

\section{Experiments and Results}
\label{Results}
To assess the QNBM model, we utilize the publicly available datasets recently curated by \cite{ALIYON2024132877} and further expanded in \cite{NBMLSS}. These datasets consist of samples from the ENTSO-E transparency platform, encompassing multiple European electricity markets. For this study, we focus on the Germany and Belgium regions, to address diverse market conditions. Exploration of additional power markets is left for future research.
The datasets span up to September 30, 2024, enabling evaluation under the heightened volatility characteristic of recent forecasting environments. Each dataset includes hourly forecasts for load and renewable generation (wind and solar) as exogenous inputs. Additionally, temporal features such as the age of the time series and day-of-week indicators are encoded using cyclical sine and cosine transformations.

The experimental setup follows the protocol adopted in the NBMLSS study \cite{NBMLSS} for coherence.
Input-output pairs are generated using a standard sliding window approach where, for each day $d$, the conditioning vector $x_d$ incorporates historical price values from days $d$–1, $d$–2, and $d$–7, along with the corresponding hour-wise day-ahead exogenous variables.

Out-of-sample testing uses data from October 1, 2023, to September 30, 2024. The initial training and validation sets (prior to recalibration) include samples from January 1, 2019, up to the start of the test period. Hyperparameter tuning is performed using one year of data preceding each test set, with Optuna \cite{optuna} employed for cross-validation, and early stopping based on a 20\% random subsample.
To manage computational demands across varying experimental setups, models are recalibrated weekly during the test phase and retrained using four sequential folds in the cross-validation process. Training is carried out using the Adam optimizer, with a maximum of 800 epochs and early stopping triggered after 20 epochs of no improvement, with a sliding window batch size of 128, structured in sub-blocks of 32 samples.

As baselines, beyond the NBMLSS architecture and the distributional neural network from \cite{NBMLSS} - both mapping Johnson's SU densities (with the latter labeled J-DNN hereafter) - we developed a quantile regression neural network (QR-DNN) following \cite{brusaferri2024onlineconformalizedneuralnetworks}. Both the QR-DNN and QNBM have been configured to provide the conditional percentiles (i.e., $|\Gamma|=99$) for each prediction step in a unique pass. All models are implemented with the TensorFlow Probability library \cite{dillon2017tensorflow}, through the development of custom layers and loss functions.
To be consistent with the J-DNN, the QR-DNN has been structured with two hidden layers, incorporating ReLU activations and dropout regularization. The number of units in each layer, the dropout rate, and the learning rate were tuned by grid search using the discrete sets [64, 128, 512, 640, 768], [0, 0.1, 0.3, 0.5], and [1e-3, 5e-4, 1e-4, 5e-5], respectively.
To align with the NBMLSS configurations, the number of units in the shared feature network of the QNMB model was tuned within a reduced range $n_u, n_z$ = [32, 64, 128]. 
For the experiments in this study, the rank parameter was set to $r = 16$. The hyperparameters selected by the procedure for the different markets are displayed in Table~\ref{hyper_tune}. Both the QR-DNN and the QNMB models obtained configurations consistent to their respective distributional setups.
Then, each model was recalibrated five times, starting from distinct random initializations. The forecasts were aggregated using standard quantile averaging. For consistency, we employed the results from \cite{NBMLSS} for the J-DNN and NBMLSS models including the RevIn and vincentization components.

\begin{table}[t!]
\caption{Tuned hyperparameters}
\label{hyper_tune}
\begin{center}
\small
\begin{tabular}{lllll}
& \multicolumn{2}{c}{QR-DNN} & \multicolumn{2}{c}{QNBM} \\
\hline
\bf{} &DE &BE &DE &BE 
\\ \hline
${n_u}$ &640   &128   &64   &64 \\ 
${l_r}$ &1e-4  &5e-4   &5e-4   &5e-4 \\
${d_r}$ &0.1   &0.1   &0.1   &0.3 \\ \hline
\end{tabular}
\end{center}
\end{table}
\renewcommand{\arraystretch}{1.2}
\begin{table}[t!]
\caption{Test set results}
\label{Test_perf}
\begin{center}
\begin{tabular}{lllllr}
\hline
\bf{DE} &PICP$_{50\%}$ &PICP$_{90\%}$ &PICP$_{98\%}$ &MAE &CRPS \\ \hline 
J-DNN    &54.3(13) &91.4(17) &97.8(23) &10.499 &3.809 \\
QR-DNN    &40.6(1) &83.1(2) &95.7(13) &10.629 &3.858\\
NBMLSS &53.8(18) &91.3(24) &97.9(23) &10.230 &3.728 \\
QNBM    &52.2(18) &91.1(23) &98.1(24) &10.411 &3.789\\ \hline
\bf{BE} &PICP$_{50\%}$ &PICP$_{90\%}$ &PICP$_{98\%}$ &MAE &CRPS \\ \hline 
J-DNN    &48.5(20) &89.0(21) &97.0(19) &13.431 &4.847\\
QR-DNN    &40.7(0) &84.4(4) &96.4(15) &13.432 &4.863\\
NBMLSS &48.7(21) &87.9(19) &97.1(20) &12.758 &4.644\\
QNBM    &47.3(18) &88.9(19) &97.9(23) &12.826 &4.653\\
\end{tabular}
\end{center}
\end{table}
\begin{figure}[t!]
\begin{center}
\includegraphics[width=1.0\linewidth]{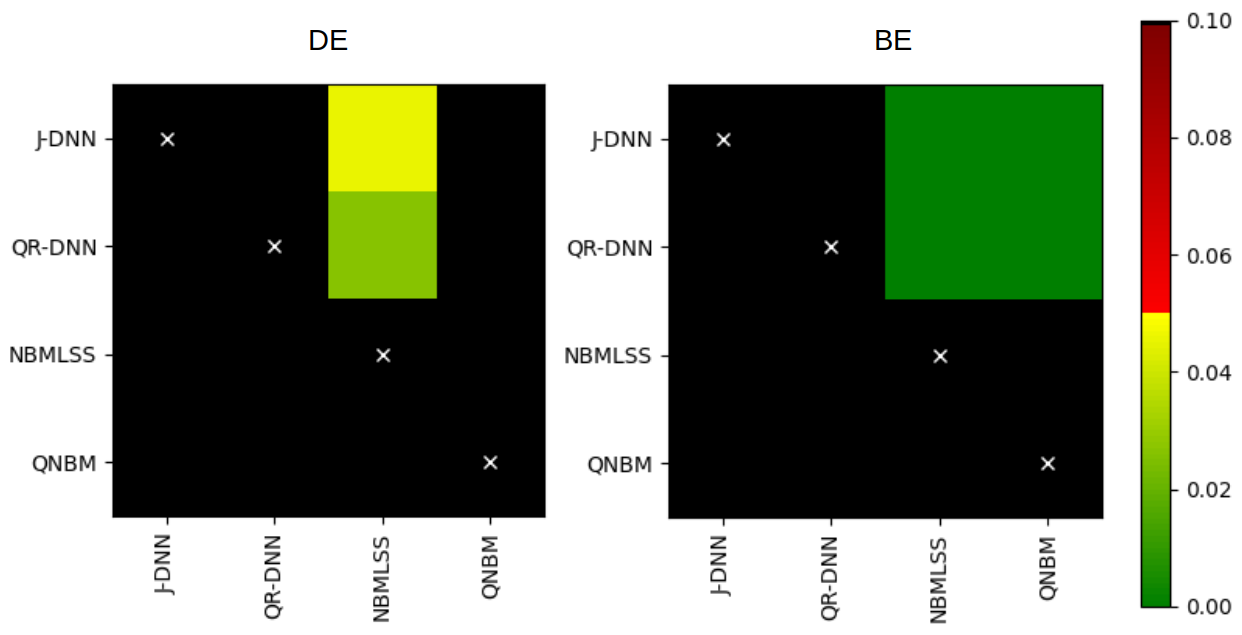}
\end{center}
\caption{DM test on test set CRPS scores}
\label{DM_Pinball_DEBE}
\end{figure}

\begin{figure}[t!]
\begin{center}
\includegraphics[width=0.95\linewidth]{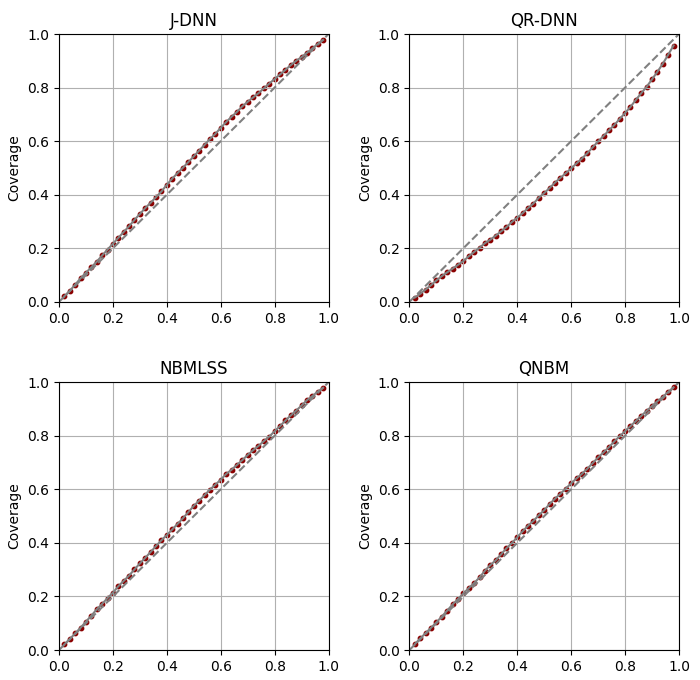}
\end{center}
\caption{DE calibration plots}
\label{DE_cali_plots}
\end{figure}

\begin{figure}[t!]
\begin{center}
\includegraphics[width=0.95\linewidth]{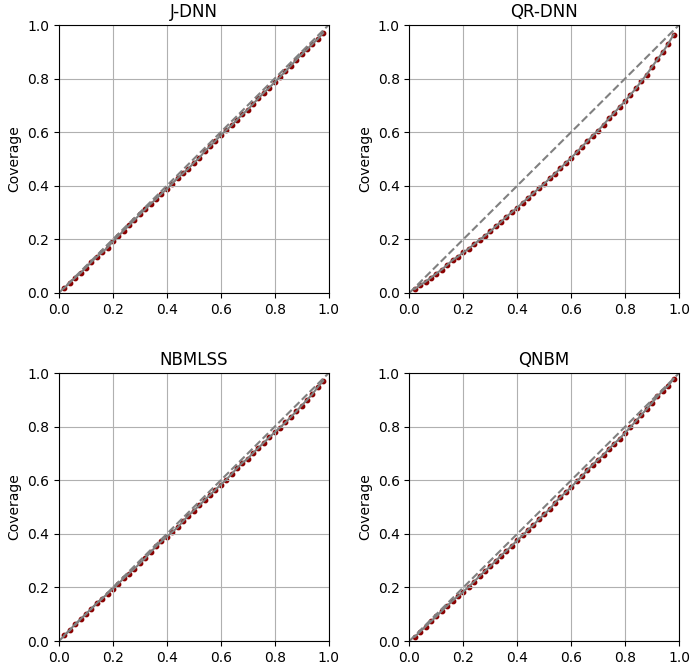}
\end{center}
\caption{BE calibration plots}
\label{BE_cali_plots}
\end{figure}

\begin{figure*}[t!]
\begin{center}
\includegraphics[width=1.00\linewidth]{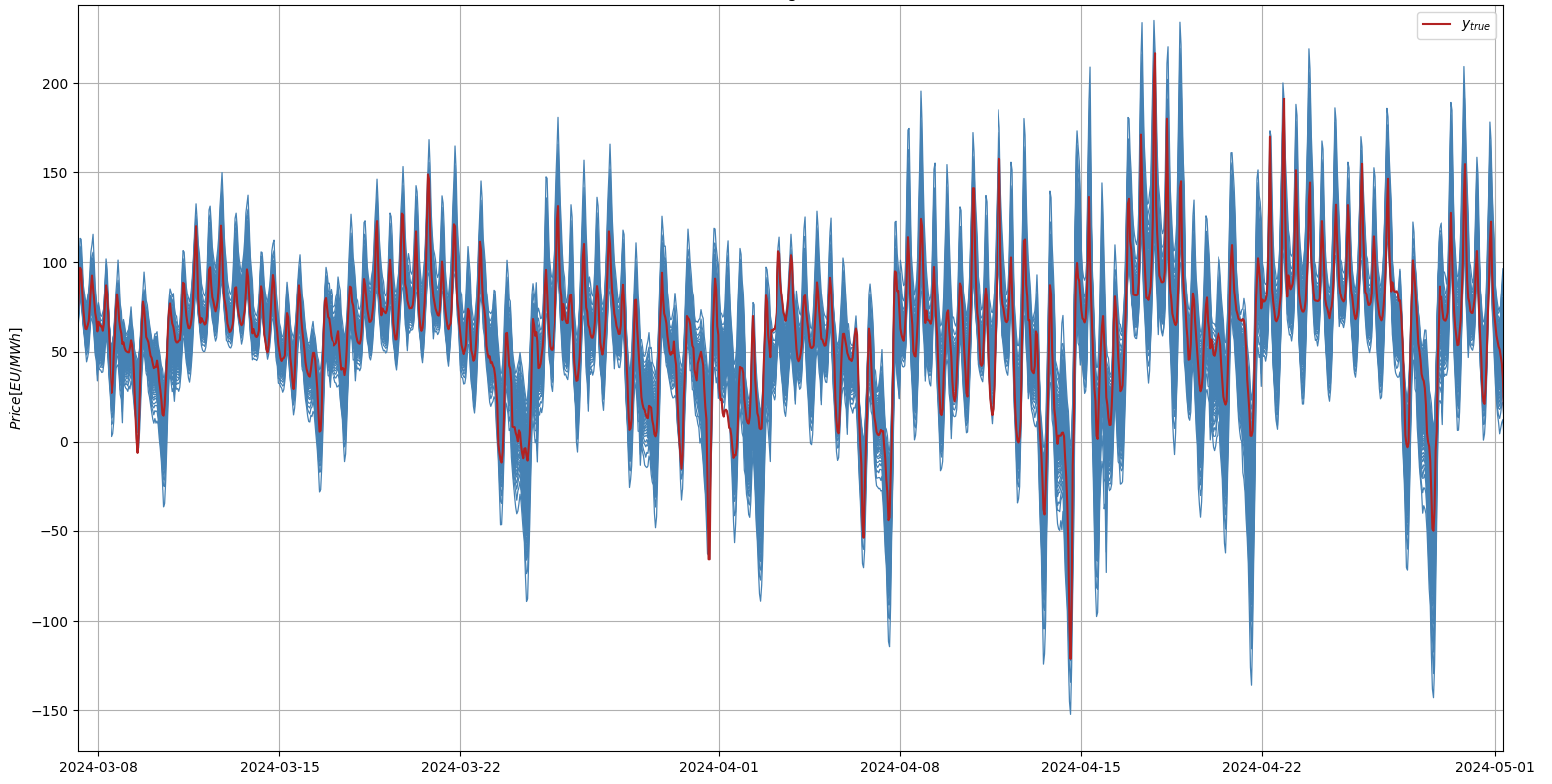}
\end{center}
\caption{Samples of predicted distribution percentiles on DE market test set}
\label{Sample_test_preds_DE}
\end{figure*}

The probabilistic forecasting results on the test sets are presented in Table~\ref{Test_perf}.
The Continuous Ranked Probability Score (CRPS) is approximated using the average Pinball loss computed across the 99 percentiles. We report the results of the Kupiec test for unconditional coverage at the 5\% significance level for both the 50\% and 90\% prediction intervals, alongside the Prediction Interval Coverage Probability (PICP). Figure~\ref{DM_Pinball_DEBE} displays the outcomes of the multivariate Diebold-Mariano (DM) test, comparing CRPS loss norms across model predictions. Additionally, the overall coverage performance across percentiles is illustrated in Figures~\ref{DE_cali_plots}–\ref{BE_cali_plots}.  
The QR-DNN baseline achieved CRPS and MAE metrics comparable to those of the distributional model, but exhibited more overconfident prediction intervals, particularly in the central quantiles, as evidenced by the coverage differences at the 50\% and 98\% levels. This behavior may be attributed to potential quantile overfitting issues, to which QR-DNN models are prone. Further investigation is warranted, for instance, by exploring additional regularization techniques. The QNBM model shows performances aligned to the NBMLSS architecture both in terms of scores and calibration. Furthermore, we observe that both the QR-DNN and the QNBM models do not exhibit the excessively large tail spikes—predominantly near zero and negative price settlements—seen in the distributional models. Samples of predicted quantiles from the DE test set are reported in Figure~\ref{Sample_test_preds_DE}. This behavior may be attributed to their more granular control over the placement of quantiles near the extremes enabled by the nonparametric setup, in contrast to the more constrained form imposed by distributional assumptions.


\begin{figure}[t!]
\begin{center}
\includegraphics[width=1.00\linewidth]{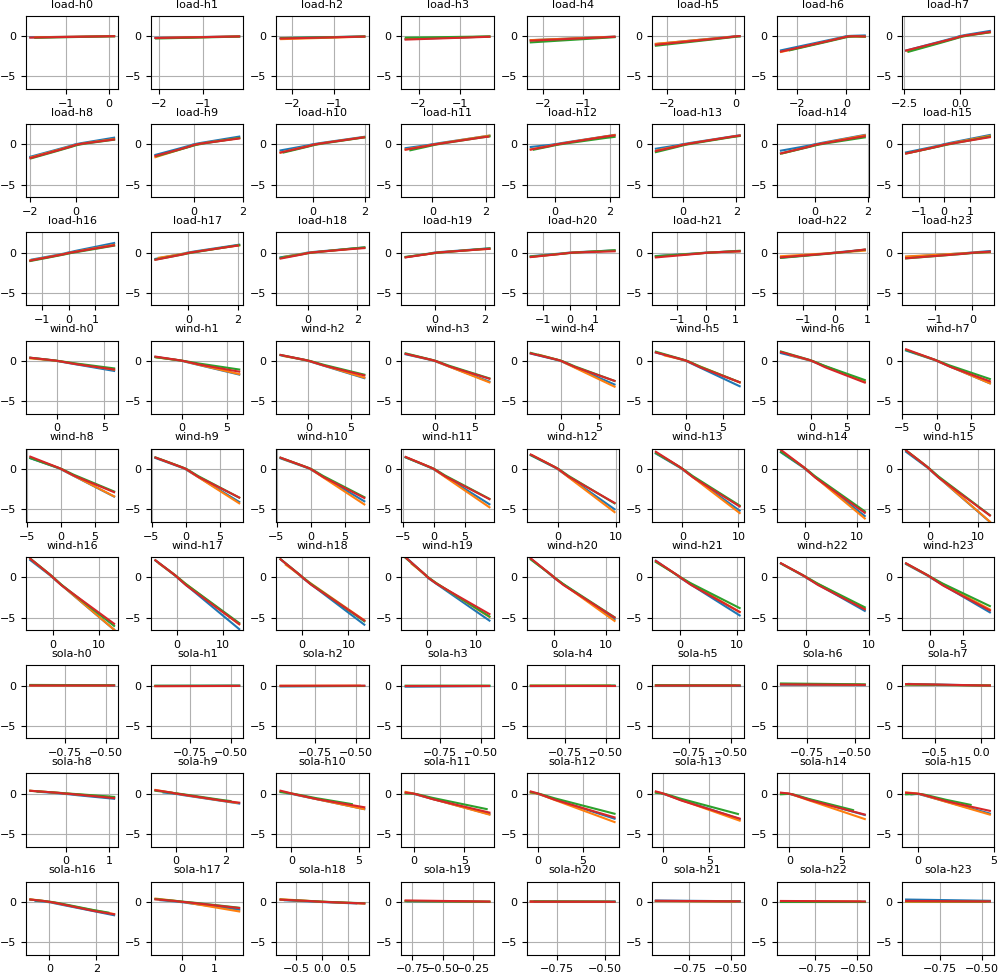}
\end{center}
\caption{Shape functions of the predicted quantile $\gamma=0.05$ on DE recalibration (5 ensemble components)}
\label{QNBM_q05}
\end{figure}

\begin{figure}[t!]
\begin{center}
\includegraphics[width=1.00\linewidth]{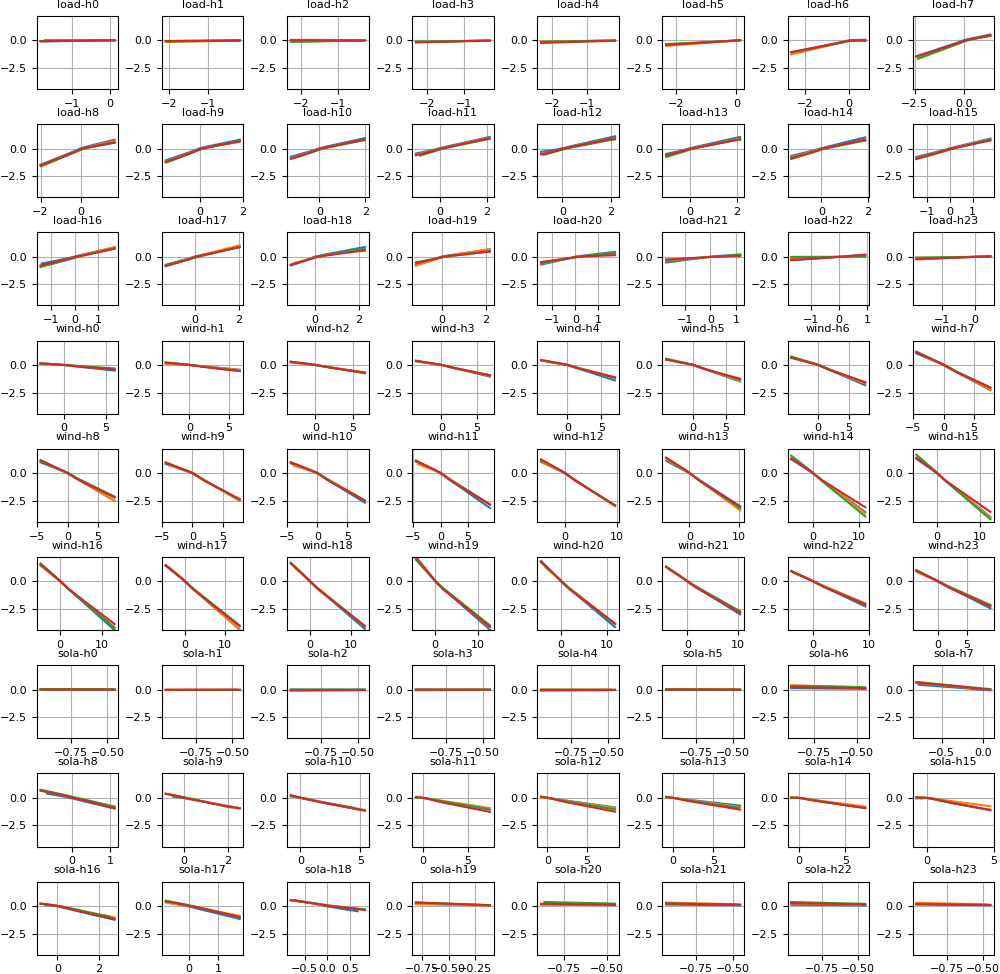}
\end{center}
\caption{Shape functions of the predicted quantile $\gamma=0.95$ on DE recalibration (5 ensemble components)}
\label{QNBM_q95}
\end{figure}

Beyond its probabilistic forecasting capabilities, the primary objective of the QNBM architecture, akin to the NBMLSS model, is to furnish users with an interpretable depiction of how each input variable influences the predicted distributional characteristics over the horizon, thus shedding light on the key factors driving the model's output.
Figures \ref{QNBM_q05}–\ref{QNBM_q95} illustrate shape functions extracted for exogenous variables influencing the predicte quantiles $\gamma=0.05$ and $\gamma=0.95$, across the horizon, on the DE market. The extraction follows the masked exogenous subsets schema applied to the final test samples of the NBMLSS study, ensuring consistency. Different colors indicate distinct ensemble components, each resulting from a separate recalibration run.
Apparently, the learning pipeline consistently converged to akin shape functions across multiple executions, showing relationships comparable to those observed in the NBMLSS model (see \cite{NBMLSS}). These include the diminishing influence of renewable generation forecasts and steeper effects of load features during peak hours compared to early morning and late evening. Nevertheless, the QNBM architecture inherits potential approximate concurvity among features (see, e.g., \cite{10.5555/3666122.3668598}, \cite{10.1007/s00180-022-01292-7}) and the general under-specification issues common in modern machine learning pipelines (see, e.g., \cite{10.5555/3586589.3586815}, \cite{10.5555/3666122.3666956}, \cite{Zhang24} and references therein). As also noted in \cite{NBMLSS}, nonlinear interactions among shape functions have not yet been thoroughly investigated in the literature. We therefore echo the recommendation of \cite{Zhang24} that effective feature selection should involve inputs from domain experts. Additionally, rather than relying on a single solution, adopting an ensemble of candidate models currently represents the most pragmatic compromise. This consideration is particularly important in forecasting contexts, where redundant information within the conditioning variables is more the norm than the exception - for instance, correlations among price values during early morning hours.
Within this framework, neural additive models such as NBMLSS and QNBM can complement the flexibility of D-DNNs and QR-DNNs, providing more transparent interpretations of feature effects throughout the input domain and thereby assisting users in the crucial stages of model development.

\section{Conclusions and Next Developments}
In this work, we have presented the Quantile Neural Basis Model (QNBM), offering a more interpretable counterpart to quantile regression neural networks (QR-DNN) through explicit shape function representations, analogous to the role Neural Basis Models for Location, Scale, and Shape (NBMLSS) play for distributional neural networks (D-DNN). By avoiding parametric distributional assumptions, QNBM enables more flexible representations of specific conditional quantiles. In contrast, NBMLSSs model the entire conditional distribution parametrically, which can limit their ability to capture complex quantile behaviors when the chosen parametric family is misspecified. However, transitioning from parameter-level to quantile-level (e.g., percentile) mappings introduces computational challenges, particularly with increasing conditioning dimensions and multi-horizon setups. To address this, we introduce matrix factorization within the basis and shape function projections beyond the shared neural network backbone.
The proposed approach has been applied to the challenging task of day-ahead electricity price forecasting in the German and Belgian power markets, covering recent periods characterized by high volatility. The experimental setup follows that of the NBMLSS study to ensure consistency and comparability. The QNBM model demonstrated performance comparable to the NBMLSS architecture while alleviating the excessive negative spike issue observed in the distributional formulation.
Overall, the distributional and nonparametric modeling approaches, as exemplified by the D-DNN/NBMLSS and QR-DNN/QNBM models respectively, represent complementary alternatives. Their deployment or integration in practice should be guided by the specific requirements of the application, balancing interpretability, flexibility, and the characteristics of the underlying data.

Several directions for future research remain open. We plan to deepen the investigation of the concurvity issue, for instance by leveraging Rashomon set approximation techniques and exploring automatic feature selection methods aimed at controlling multivariate concurvity. Additionally, we intend to explore the incorporation of further regularization strategies and architectural enhancements, alongside extending our investigation to a broader range of forecasting applications.

\bibliography{mybibfile_v01}

\end{document}